%% file: arxiv_submission.tex
\documentclass[10pt,twocolumn,letterpaper]{article}

\usepackage{algorithm}

\usepackage{microtype}
\usepackage{graphicx}
\usepackage{latexsym,amssymb,amstext,amsfonts,amsmath,setspace,bbm,mathtools}
\usepackage{subfig}
\usepackage{xcolor}
\usepackage[noend]{algpseudocode}
\usepackage{adjustbox}
\usepackage{amsthm}

\usepackage[font={small,it}]{caption}

\DeclareMathOperator*{\argmax}{arg\,max}
\DeclareMathOperator*{\mL}{\mathcal{L}}

\newcommand{\tb}[1]{\textbf{#1}}
\newcommand{\yh}{\hat{y}}
\newcommand{\xh}{\hat{x}}
\newcommand{\Tch}{\hat{\mathbb{T}}}
\newcommand{\Tc}{\mathbb{T}}

\newcommand{\thetahat}{\hat{\theta}}
\newtheorem{proposition}{Proposition}
\newtheorem{lemma}{Lemma}
\newcommand{\ignore}[1]{}

\theoremstyle{definition}
\newtheorem{definition}{Definition}[section]

\newcommand{\bbE}{\mathbb{E}}
\newcommand{\one}{\mathbf{1}}

\DeclarePairedDelimiter\floor{\lfloor}{\rfloor}

\usepackage{longtable}

\usepackage{geometry}
\usepackage[pagebackref=true,breaklinks=true,colorlinks,bookmarks=false,colorlinks = true,
            linkcolor = blue,
            urlcolor  = blue,
            citecolor = blue,
            anchorcolor = blue]{hyperref}

\def\citep{\cite}
\def\citet{\cite}

\newcommand{\comment}[1]{}

\makeatletter
\renewcommand*{\@fnsymbol}[1]{\ensuremath{\ifcase#1\or *\or \dagger\or \ddagger\or
    \mathsection\or *\or \|\or **\or \dagger\dagger
    \or \ddagger\ddagger \or 1 \else\@ctrerr\fi}}
\makeatother

 \title{Constrained Classification and Ranking via Quantiles}

\author{
    Alan Mackey
    \thanks{Google AI Perception}\\
    {\tt\small mackeya@google.com}
    \and 
    Xiyang Luo 
    \thanks{Department of Mathematics, UCLA}\\
    {\tt\small mathluo@math.ucla.edu}
    \and
    Elad Eban
    \footnotemark[1]\\
    {\tt\small elade@google.com}}

\begin{document}
\setcounter{footnote}{0}
\renewcommand{\thefootnote}{\alph{footnote}}

\maketitle
 \input{content}
 \bibliography{references.bib}
 \bibliographystyle{plain}


\appendix
\newpage
\input{proofs.tex}

\end{document}

%% file: content.tex
\begin{abstract} 
In most machine learning applications, classification accuracy is not the primary metric of interest. Binary classifiers which face class imbalance are often evaluated by the $F_\beta$ score, area under the precision-recall curve, Precision at K, and more. The maximization of many of these metrics can be expressed as a constrained optimization problem, where the constraint is a function of the classifier's predictions.

In this paper we propose a novel framework for learning with constraints that can be expressed as a predicted positive rate (or negative rate) on a subset of the training data. We explicitly model the threshold at which a classifier must operate to satisfy the constraint, yielding a surrogate loss function which avoids the complexity of constrained optimization. 
The method is model-agnostic and only marginally more expensive than minimization of the unconstrained loss. Experiments on a variety of benchmarks show competitive performance relative to existing baselines.
\end{abstract}

\section{Introduction}
%
%
In many real-world applications, machine learning classification systems do not simply make predictions to minimize errors or to maximize likelihood. Instead, the decision threshold of a learned classifier is often adjusted after training to select a particular operating point on the precision-recall or ROC curve, reflecting how the classifier will be used. Automated email spam filters, for example, might operate with an increased threshold to achieve a high precision or low predicted positive rate. In medical diagnosis, the threshold may be decreased so that the classifier will make predictions with a high recall \cite{he2009learning}. When a particular operating point for the classifier is not known ahead of time, area under the precision-recall or ROC curve is often the metric used to compare models. In other cases, classifiers are adjusted to obey constraints on coverage or fairness (i.e. predicted positive rates on subsets of the training data) so that the system will not treat some demographics of users unfairly \cite{goh2016satisfying}. 

In all of these cases, the most desirable binary classifier can be characterized as one which maximizes a quantity such as accuracy, precision, or recall, subject to a constraint. The constraint is usually an inequality on the predicted positive rate (coverage) or true/false positive/negative rate on some subset of the data. The most common practice to produce a classifier which satisfies a constraint is to train the classifier by maximum likelihood, then after training, adjust its threshold so that the constraint is satisfied.

Threshold adjustment enjoys a strong theoretical justification \cite{koyejo2014consistent}: for a large family of metrics, the Bayes optimal classifier makes predictions by thresholding the conditional class probability $P(y=1|x)$. The same result is known for binary cost-sensitive classification \cite{elkan2001foundations}. However, thresholding is not the end of the story because learning accurate (or calibrated) conditional probabilities is more challenging than learning loss-minimizing predictions \cite{joachims2005support, margineantu2002class}. This observation is a fundamental motivation for structured losses. 

Accordingly, there have been several attempts to address constrained classification problems directly through the machinery of constrained optimization \cite{davenport2010tuning,goh2016satisfying,eban2017scalable}. Instead, we focus on eliminating the constraint by substitution. For constraints which can be expressed as a rate on the training dataset (e.g. predicted positive rate, the threshold of a classifier which satisfies the constraint can be expressed as a quantile of the classifier's scores. By incorporating an estimator of the quantile into the loss function, the constraint can be eliminated. For simplicity, we focus on binary classification and apply the quantile estimation approach to optimize precision at a fixed recall and precision@K.

In summary, our contribution is as follows. First, we show that a wide variety of machine learning problems with constraints can be recast as ordinary, unconstrained learning problems. Second, we show that the resulting unconstrained problems can be solved efficiently and generally; they are particularly amenable to gradient-based optimization, independent of what model is used. For linear models, we derive a convex upper bound on the loss, and also prove a uniform risk bound. Finally, in experiments across a variety of benchmarks, we demonstrate performance matching or outperforming state of the art methods.

\section{Related Work}
%
%
Many existing approaches aim to optimize for non-decomposable performance measures, enforce dataset constraints, or incorporate classifier score quantiles into the training procedure. Below, we summarize those most related to the proposed method. 

Several attempts, most notably \cite{joachims2005support}, have used the framework of structured prediction \cite{taskar2004max, tsochantaridis2005large, koller2009probabilistic} to handle performance metrics beyond accuracy. The method, however, faces several problems. First, each iteration of the algorithm in \cite{joachims2005support} requires fully training an SVM over the entire dataset, which hinders scalability. Second, it is not clear how the methods extends to handle constraints which depend on the labels, which are not available at test time. For example, a constraint on the predicted positive rate is easily enforced at test time, while a constraint on the true positive rate is not. 

The authors of \cite{kar2015surrogate} identify a third shortcoming: the structured SVM framework \cite{joachims2005support} assumes that the true labels are admissible under the constraint on the classifier. When optimizing Precision@K, the true labels are only admissible when the dataset contains exactly $K$ positives. This problem and the scalability issue are both addressed in \cite{kar2015surrogate}, but how to handle constraints which depend on the ground truth labels remains unclear. 

For the specific case of a linear classifier, \cite{goh2016satisfying} approaches general dataset constraints by framing the problem as constrained optimization with the ramp loss and solving it with a majorization-minimization approach. However, their algorithm is not readily generalized to the nonlinear case. 

The special case of minimizing the probability of false negatives subject to a constraint on the probability of false positives is known as Neyman-Pearson classification \cite{tong2016survey}. It is known that, with minor regularity assumptions, Neyman-Pearson classification can be accomplished by re-weighting the losses on the positives and negatives. Both \cite{davenport2010tuning} and \cite{eban2017scalable} seek to identify the appropriate re-weighting during the course of training, the former by direct search and the latter by saddle-point optimization with Lagrange multipliers. These methods apply to non-linear classifiers such as neural networks, and \cite{eban2017scalable} extends the idea to handle the $F_\beta$ meaure, ROC-AUC, and Precision-Recall AUC.

The saddle-point approach has an advantage when direct search is very expensive and it applies to a wider class of models, but there is no guarantee that the constraint will be satisfied unless the saddle point problem is solved to convergence. Unfortunately, this is incompatible with the ubiquitous practice of early stopping in neural networks. In addition, the convex relaxation of the constraint can result in loose bounds and oversatisfied constraints.

The idea of placing a quantile constraint on the classifier's scores was explored in \cite{boyd2012accuracy}, and produced impressive results relative to other approaches for optimizing ranking metrics such as precision@K. However, the method of \cite{boyd2012accuracy} requires learning a separate classifier for each datapoint in the training set, which is tractable only for very small datasets. In contrast, our approach is amenable to stochastic gradient methods and scales easily to datasets with millions of examples.

\section{Quantile Surrogate Losses}
%
%
In this section, we describe our approach in detail and focus on maximizing Precision@K and Precision at a fixed recall. 
Let $X = \{x_1, \dots, x_n\}$ consist of the features in a training dataset, and let $Y = \{y_1, \dots, y_n\}$ denote the corresponding labels in $\{-1, 1\}$.

We consider a classifier to consist of a scoring function $f$ determined by learnable parameters $w$, and a threshold $\theta$. 
The prediction $\yh_i$ of the classifier on the $i$th example is made according to 
\[
    \yh_i = \text{sign}(f(x_i; w) - \theta).
\]

\begin{definition}
Given a subset $A \subseteq X$, we define the \emph{predicted positive rate} for a classifier $(f, \theta)$ on $A$ to be the fraction of examples in $A$ predicted to be positives:
$$r_A(f, \theta) = \frac{|\{x_i \in A: f(x_i; w) - \theta > 0\}|}{|A|}.$$
A \emph{rate constraint} is a constraint which can be written in the form $r_A \gtrless c$
for some value of $c\in [0, 1]$.
\end{definition}

Note that $r_A(f, \theta)$ is piecewise constant, discontinuous, and monotone decreasing in $\theta$. Defining $f(A) = \{f(x): x \in A\}$, we see that $r_A(f, \min f(A)) = 1$ and $r_A(f, \max f(A)) = 0$. Therefore, inequality rate constraints can always be satisfied by setting $\theta$ to some value depending on $f$ and $c$.

The notion of predicted positive rate on a subset is general enough to represent many metrics of interest. Letting $X^+$ denote the positive examples in $X$, we see that the recall (or true positive rate) of a classifier $(f, \theta)$ is equal to $r_{X^+}(f, \theta)$. The typical notion of predicted positive rate coincides with $r_X(f, \theta)$, and false positive rate can be written as $r_{X^-}(f, \theta)$. Other examples are given in \cite{goh2016satisfying}: coverage, churn, and fairness can all be expressed in terms of predicted positive rates. Because the predicted positive and predicted negative rates must sum to one, we may consider predicted positive rate constraints without loss of generality. 

Given some metric (or utility) $G$ that depends on the data $X, Y$ and the classifier $(f, \theta)$, we can define the problem of maximizing $G$ subject to the rate constraint $r_A(f, \theta) \geq c$ as
\begin{equation}\label{eq:constrained_optimization}
    \max_{f, \theta} G(X, Y, f, \theta) \quad \text{subject to} \quad r_A(f, \theta) \geq c.
\end{equation}
Note that $G$ need not decompose across examples, as accuracy does. Due to the discontinuity of $r_A(f, \theta)$ and most metrics $G$, the problem is combinatorial and intractable to solve exactly. However, there are many cases in which we can characterize an optimizer $(f^*, \theta^*)$ in terms of the quantiles of $f^*(A)$.

\begin{definition}
For a set of  real numbers $S = \{s_1, \ldots, s_N\}$ the quantile function is defined as
\begin{equation*}
\begin{split}
    q(S, c) &= \sup \{t: |\{s_i \leq t\}|/N \leq c\} \\
            &= \inf \ \{t: |\{s_i \leq t\}|/N \geq c\}.
\end{split}
\end{equation*}
\end{definition}

\begin{proposition}\label{prop:quantile}
Suppose $(f, \theta)$ is a feasible point for (\ref{eq:constrained_optimization}) and let
\[
    \hat{\theta} = q(f(A), 1 - c).
\]
Then $(f, \hat{\theta})$ is also feasible. If $G(X, Y, f, \theta)$ is monotone increasing in $\theta$ over the range $[\theta, \hat{\theta}]$, then $G(X,Y,f,\hat{\theta}) \geq G(X,Y,f,\theta)$.
\end{proposition}
\begin{proof}
Because $r_A(f, \theta)$ is monotone decreasing in $\theta$, any admissible $\theta$ must satisfy
\begin{equation*}
\begin{split}
    \theta &\leq \sup \{t: r_A(f^*, t) \geq c\} \\
           &\leq \sup \{t: |\{x \in A : f^*(x) > t\}|/|A| \geq c\} \\
           &\leq \sup \{t: 1 - |\{x \in A: f^*(x) \leq t\}|/|A| \geq c\} \\
           &\leq \sup \{t: |\{x \in A: f^*(x) \leq t\}|/|A| \leq 1-c \}.
\end{split}
\end{equation*}

The supremum in the last line is exactly the quantile function, and so 
$$\theta \leq q(f(A), 1 - c).$$
The claim follows from the fact that $G$ is monotone in $\theta$ and that $\hat{\theta} \geq \theta$.
\end{proof}

Applying the proposition to an optimizer $(f^*, \theta^*)$ shows that $(f^*, \hat{\theta})$ is also an optimizer. In particular, a scoring function $f$ which optimizes (\ref{eq:constrained_optimization}) is a minimizer of the loss
\begin{equation}\label{eq:unconstrained_optimization}
    \min_f L(X, Y, f, q(f(A), 1-c))
\end{equation}
With $L = -G$. Conversely, optimizers of (\ref{eq:unconstrained_optimization}) are feasible and hence optimal for (\ref{eq:constrained_optimization}).
In practice, $G$ is usually a differentiable surrogate allowing for numerical optimization and the monotonicity assumption is easily verified. For inequality constraints of the form $r_A(f, \theta) \leq c$, $\theta^*$ is again given by the quantile function so long as $G$ is monotone decreasing in $\theta$. 


Proceeding 
farther depends on the choice of $G$ and the specific rate constraint, so we consider the task of maximizing precision subject to recall and predicted positive rate constraints. Afterward, we consider the estimation of the quantile function $q(f(A), c)$.

\subsection{Precision at a fixed recall}
%
%
The precision and recall of a classifier $(f, \theta)$ are defined as
\begin{equation*}
\begin{split}
      P(f, \theta) &= \frac{TP(f, \theta)}{TP(f, \theta) + FP(f, \theta)} \\
      R(f, \theta) &= \frac{TP(f, \theta)}{TP(f, \theta) + FN(f, \theta)} = \frac{TP(f, \theta)}{|X^+|},
\end{split}
\end{equation*}
where $TP$ and $FP$ denote the true positives and false positives, respectively:
\begin{equation*}
  \begin{split}
      TP(f, \theta) &= \sum_{i: y_i = 1} \mathbf{1}_{f(x_i; w) > \theta}\\
      FP(f, \theta) &= \sum_{i: y_i = -1} \mathbf{1}_{f(x_i; w) > \theta}.
  \end{split}
\end{equation*}

Thus, to optimize for the Precision@Recall objective (P@R), we wish to solve
\begin{equation}\label{eq:max_p_at_r}
\begin{split}
    & \max_{f, \theta} \frac{TP(f, \theta)}{TP(f, \theta) + FP(f, \theta)} \\
    & \text{subject to} \quad r_{X^+}(f, \theta) \geq c.
\end{split}
\end{equation}

We know the recall constraint $r_{X^+} \geq c$ will be active, because precision is trivially maximized by predicting few or no positives. This implies $r_{X^+} = \tilde{c}$, where $\tilde{c} = \min \{k/|X^+|: k/|X^+| \geq c\}$. Because $N^+ = |X^+|$ is fixed, we conclude that $TP(f, \theta) = \tilde{c} N^+.$ Substituting this value for $TP$ into the expression for precision gives
\[
    P(f, \theta) = \frac{TP(f, \theta)}{TP(f, \theta) + FP(f, \theta)} = \frac{cN^+}{cN^+ + FP(f, \theta)}.
\]

Thus, solving
\begin{equation}\label{eq:p_at_r_lb}
\begin{split}
    & \max_{f, \theta} \frac{\tilde{c} N^+}{\tilde{c} N^+ + FP(f, \theta)} \\ 
    & \text{subject to} \quad r_{X^+}(f, \theta) \geq c
\end{split}
\end{equation}
gives the solution to (\ref{eq:max_p_at_r}). In addition, (\ref{eq:p_at_r_lb}) is equivalent to
\begin{equation}\label{eq:min_fp_at_r}
    \min_{f, \theta} FP(f, \theta) \quad \text{subject to} \quad r_{X^+}(f, \theta) \geq c.
\end{equation}
With the objective in this form, $FP(f, \theta)$ can be upper bounded in the standard way by the logistic loss (or hinge loss), which we denote by $l$:
\begin{equation*}
\begin{split}
      FP(f, \theta) &= \sum_{i: y_i = -1} 1_{f(x_i; w) - \theta > 0} \\
                    &\leq \sum_{i: y_i = -1} l(f(x_i; w) - \theta).
\end{split}
\end{equation*}
This leaves us with
\begin{equation}\label{eq:min_bound_at_r}
\begin{split}
    & \min_{f, \theta} \sum_{i: y_i = -1} l(f(x_i; w) - \theta) \\
    & \text{subject to} \quad r_{X^+}(f, \theta) \geq c.
\end{split}
\end{equation}
Because the loss is monotone decreasing in $\theta$, the monotonicity assumption of Proposition \ref{prop:quantile} is met and so $\theta = q(f(X^+), 1-c)$. This leads to the unconstrained objective for P@R:
\begin{equation}\label{eq:unconstrained_fp_at_r}
    \min_f \sum_{i: y_i = -1} l(f(x_i; w) - q(f(X^+), 1-c)).
\end{equation}
In practice, the minimization is performed with respect to the parameters $w$ of the scoring function $f$, e.g. by stochastic gradient descent. 

As pointed out in \cite{eban2017scalable}, the precision at recall objective can be used to target Precision-Recall AUC by approximating the area under the Precision-Recall curve as a Riemann sum.

\subsection{Precision a fixed predicted positive rate} \label{sec:ppr}
%
%
For a training dataset $X$ with $N$ points, the Precision@K metric is equivalent to precision at a predicted positive rate of $K/N$. Therefore, we consider the objective
\begin{equation} \label{eq:max_p_at_ppr}
    \max_{f, \theta} P(f, \theta) \quad \text{subject to} \quad r_X(f, \theta) = c.
\end{equation}

Because $TP + FP$ is equal to the number of predicted positives, the constraint $r_X(f, \theta) = c$ implies that $TP + FP = cN$, and so (\ref{eq:max_p_at_ppr}) is equivalent to 
\[
    \max_{f, \theta} TP(f, \theta) \quad \text{subject to} \quad r_X(f, \theta) = c.
\]
Using $TP = cN - FP$, the objective can be rewritten as
\[
    \min_{f, \theta} FP(f, \theta) \quad \text{subject to} \quad r_X(f, \theta) = c,
\]
which by the same logistic loss bound and application of Proposition \ref{prop:quantile} becomes
\begin{equation}\label{eq:unconstrained_fp_at_ppr}
    \min_f \sum_{i: y_i = -1} l(f(x_i; w) - q(f(X), 1 - c)).
\end{equation}

Alternatively, because the loss on the positive examples can be used as a lower bound on the true positive rate \cite{eban2017scalable}, we can consider the objective
\begin{equation}\label{eq:unconstrained_tp_at_ppr}
    \min_f \sum_{i: y_i = 1} l(-f(x_i; w) + q(f(X), 1 - c)).
\end{equation}

\section{Estimating the Quantile Function}
%
%
Even with the unconstrained losses (\ref{eq:unconstrained_fp_at_r}) and (\ref{eq:unconstrained_fp_at_ppr}) in hand, we are left with the question of how to estimate the quantile function
$$q(f(A), c).$$
Whatever estimator we choose is required to have explicit dependence on $w$, for the purposes of numerical optimization.

The simplest choice is to apply the definition of $q$ directly, which results in the point estimator 
\[
    \hat{q}_1(f(A), c) = f(\hat{x})
\]
where $\hat{x}$ is the datapoint which solves
\[
    \hat{x} = \argmax_{x \in A} \left\{ f(x): \frac{|\{z \in A: f(z) \leq f(x)\}|}{|A|} \leq c \right\}.
\]
In other words, we take the scores $f(A) = \{f_1, \ldots, f_n\}$ sorted in ascending order and use $\hat{q}_1 = f_k$ for the largest integer $k$ such that $k/n \leq c$. 

The concern with $\hat{q}_1$ is that the variance of this estimator and its gradient may be problematically large. For example, consider the case when the scoring function $f$ is linear: 
$$f(x; w) = w^Tx,$$
where a bias term is unnecessary because it can be absorbed in to $\theta$. In this case, the loss for a rate constraint $r_A(f, \theta) \geq c$ is 
\begin{equation}\label{eq:one_point_loss}
\begin{split}
    L &= \sum_i l(f(x_i; w) - \hat{q}_1(f(A))) \\
      &= \sum_i l(w^Tx_i - w^T\hat{x}),
\end{split}
\end{equation}

where $\hat{x} \in A$ is the datapoint such that $\hat{q}_1(f(A), c) = f(\hat{x}) = w^T \hat{x}$. Note that, due to the change in $\hat{x} = \hat{x}(w)$ as $w$ changes, the loss is not convex. 

Letting $\sigma(x)$, the logistic sigmoid, denote the derivative of logloss, the gradient of $L$ (where it is defined) is
\begin{equation}
\begin{split}
    \nabla_w L &= \sum_i \sigma(w^Tx_i - w^T\hat{x}) (x_i - \hat{x}) \\
               &= \sum_i a_i x_i - \left(\sum_i a_i \right) \hat{x}
\end{split}
\end{equation}
where $a_i = \sigma(w^Tx_i - w^T\hat{x})$. From this expression, the excessive influence of $\hat{x}$ is clear. Variation in the classifier parameters or the data which causes only a small change in the quantile estimate may nonetheless cause a dramatic change in $\nabla_w L$; this gradient is discontinuous. 

In \citet{boyd2012accuracy}, an objective similar to (\ref{eq:unconstrained_fp_at_ppr}) is considered in the presence of rate constraints. There, (\ref{eq:one_point_loss}) is minimized separately for each possible choice $\hat{x} \in X$ to yield a solution $w_k$ for each $k = 1, \ldots, |X|$. Of these $|X|$ solutions, the algorithm selects the one with the smallest value of $|r_A(f(\cdot; w_k), \theta) - c|$. While this approach elegantly handles the nondifferentiability and nonconvexity of $L$, it is unfortunately not feasible even for datasets of moderate size.

One alternative for estimating $q(f(A), c)$ would be to assume a flexible parametric form for the distribution of the scores $f(A)$, for which the quantile function is available in closed form (e.g. as a function of the maximum likelihood parameter estimates). While this approach might suffice for simple scoring functions, its utility is dubious in the context of complex scoring functions such as neural networks \cite{kull2017beta}. 

\subsection{Kernel Quantile Estimators}
%
%
To achieve lower gradient variance than the point estimator without relying on parametric assumptions, we turn to \textit{kernel quantile estimators}. These are a subclass of $L$-estimators, computed as a weighted average of the order statistics of $f(A)$ \cite{sheather1990kernel, zielinski2004optimal}. These estimators are promising for our application because their gradients are far less sensitive to small parameter changes than the point estimator. 

\begin{definition}
Let $S = \{s_1, \ldots, s_N\}$ be a set of real numbers, and let $s_{(i)}$ denote the $i$th order statistic of $S$ (i.e. $s_{(i)} = s_{\sigma(i)}$ where $\sigma$ is the permutation which sorts $S$ in ascending order). 
Given a symmetric, normalized kernel $\phi$ and a nonnegative scale parameter $h$, the \textit{kernel quantile estimator} corresponding to $\phi$ and $h$ is defined \cite{sheather1990kernel}
\begin{equation}\label{eq:kernel-estimator}
\begin{split}
    \hat{q}^h_\phi(S, c) = \frac{1}{N} \sum_{i=1}^N \phi_h(i^*/N - c) s_{(i)},
\end{split}
\end{equation}
where $\phi_h(x) = \frac{1}{h}\phi(x/h)$, $c$ is the quantile to be estimated, and the index $i^*$ is defined to break ties: $i^* = \max \{j: s_{(j)} = s_{(i)}\}$.
\end{definition}

The free parameter $h$ controls the scale of the kernel, and increasing it trades off variance for bias. There are many other $L$-estimators of the quantile function; see for example \cite{cheng1995bernstein, harrell1982new, kalgh1982generalized, pepelyshev2014estimation, zielinski2004optimal} and citations therein.
It is beyond the scope of this paper to consider all of them, so in all experiments we use the Gaussian kernel estimator
\[
  \phi_h(x) = \frac{1}{h \sqrt{2 \pi}} e^{x^2 / 2 h^2}.
\]
Like the point quantile estimator, the kernel estimators lead to losses which are not convex.

We also consider an $L$-estimator which serves as a lower bound for the point estimator and results in convex loss.
Let
\begin{equation}
\label{eq:convex-estimator}
\begin{split}
    \hat{q}_m(S, c) = \text{mean}_{x \in S} \left\{x: \frac{|\{z \in S: z \leq x\}|}{|S|} \leq c \right\}. 
\end{split}
\end{equation}
From the definition of the point estimator, it is immediate that $q_1 \ge q_m$ as the \emph{max} is greater then the \emph{mean}. 
In other words: rather than taking the $k$th largest score, where $k$ is the largest integer $k$ such that $k/N \leq c$, we take the mean of the bottom $k$ scores, which serves as a lower bound. In the case when $f$ is linear, this lower bound is concave because it is a pointwise minimum of affine functions \cite{boyd2004convex}. 

Returning to the case of a rate-constrained precision loss (of which (\ref{eq:unconstrained_fp_at_r}) and (\ref{eq:unconstrained_fp_at_ppr}) are special cases), we see that because $\hat{q}_m \leq \hat{q}_1$ and logloss is nondecreasing,
\begin{equation} \label{eq:one_point_bound}
\begin{split}
    & \sum_i l (f(x_i; w) - \hat{q}_1(f(A), c)) \\
    \leq & \sum_i l (f(x_i; w) - \hat{q}_m(f(A), c))
\end{split}
\end{equation}
so that using the $\hat{q}_m$ estimator yields an upper bound on the loss with the point estimator $\hat{q}_1$. When $f$ is linear, repeated applications of the rules of convex function composition \cite{boyd2004convex} show that $\hat{q}_m(f(A), c)$ is convex and hence the entire upper bound is as well. The bound $\hat{q}_m$ enjoys a lower gradient variance than $\hat{q}_1$, and is tightest when $c$ is small, which occurs exactly when enforcing a constraint that the predicted positive rate on $A$ be high. 

\subsection{Stochastic Gradient Descent with Quantile Estimators}
Losses of the form (\ref{eq:unconstrained_optimization}), regardless of the choice of quantile estimator or bound, are compatible with any scoring function and amenable to numerical optimization. For concreteness we consider stochastic gradient descent, described in Algorithm \ref{euclid}.

\begin{algorithm}
\caption{SGD for Quantile loss (\ref{eq:unconstrained_optimization})}\label{euclid}
\begin{algorithmic}[1]
\Require A dataset $(X, Y)$, desired rate constraint $r_A \geq c$ on a subset $A \subseteq X$, quantile estimator $\hat{q}$, scoring function $f(\cdot; w)$, and learning rate $\gamma$.
\While{not converged}
  \State Gather a minibatch $(X_b, Y_b)$ from $(X, Y)$ 
  \State Gather a minibatch $A_b$ from $A$ \label{alg:step2}
  \State Update \linebreak $w \gets w - \gamma \nabla_w L(X_b, Y_b, f(\cdot; w), \hat{q}(f(A_b), 1-c))$
\EndWhile
\end{algorithmic}
\end{algorithm}

Step \ref{alg:step2} is beneficial when datapoints from $A$ are rare, but when $A$ is common or minibatches are large it will suffice to take $A_b = X_b \cap A$.

\section{Experiments}
\subsection{Precision@K}
In this section we consider the application of the quantile threshold framework to the target metric Precision@K. We compare the quantile loss with two state of the art algorithms: Accuracy at the top \cite{boyd2012accuracy} and the average surrogate loss $l^\text{avg}_{prec@k}$ from \cite{kar2015surrogate}. To target Precision@K, we follow \cite{boyd2012accuracy} and optimize for Precision at a predicted positive rate $\tau = K/N$, where $N$ is the number of datapoints. In all experiments, the Gaussian kernel quantile estimator was used.

\subsubsection{Ionosphere and Housing datasets}
We compare the results from \cite{boyd2012accuracy} to those obtained using the quantile loss surrogate, averaged across 100 random train/test splits of the data. The fraction of data used for training is 30\% for Ionosphere and 66\% for Housing; the rest is held out for testing. Tables \ref{tab:ionosphere} and \ref{tab:housing} show the Precision@$\tau$ of the methods, where $\tau$ is the classifier's predicted positive rate. The models trained with the quantile loss surrogate were evaluated at the same value of $\tau$ for which they were trained, and were optimized using gradient descent with momentum on the objective (\ref{eq:unconstrained_fp_at_ppr}) with weight decay. 

The weight decay regularization coefficient and scale parameter $h$ of the kernel quantile estimator are the algorithm's only hyperparameters. For a fair comparison against Accuracy at the top, which has only one hyperparameter (the regularization coefficient), we fix $h=0.05$ on the Ionosphere data and $h = 0.08$ on the Housing data. As in \cite{boyd2012accuracy}, for each value of $\tau$ the regularization coefficient $C$ was chosen based on the largest average value of Precision@$\tau$.

Because the quantile surrogate loss is nonconvex, the quantile method may converge to a suboptimal local minimum. To mitigate this problem, we run the algorithm with multiple random initializations of $w$ and take the solution with the lowest loss on the training set. Results for one and three initializations are reported.

The quantile surrogate achieves results matching or beating Accuracy at the top, with the largest improvements occurring for small $\tau$. In addition, optimization of the quantile surrogate enjoys very favorable computational complexity relative to Accuracy at the top. Assuming the same number of iterations across all algorithms, logistic regression has an $O(N)$ cost. Accuracy at the top requires solving a separate logistic regression problem for each datapoint, for a total cost of $O(N^2)$. On the other hand, the only additional cost of the quantile method over logistic regression is a sorting operation per iteration, for a total cost of $O(N\log N)$.

\begin{table}[!htp]
    \centering
    \begin{adjustbox}{width=0.47\textwidth}
    \begin{tabular}{|c|c|c|c|c|}\hline
        $\tau$ (\%) & LR                    & AATP                      & Q1                & Q3  \\ \hline
                  1 & 0.52 $\pm$ 0.38       & 0.85 $\pm$ 0.24         & 0.87 $\pm$ 0.27   & \tb{0.98 $\pm$ 0.10} \\ \hline
                  5 & 0.76 $\pm$ 0.14       & 0.91 $\pm$ 0.14         & 0.93 $\pm$ 0.11   & \tb{0.98 $\pm$ 0.07} \\ \hline
                9.5 & 0.83 $\pm$ 0.08       & 0.93 $\pm$ 0.06         & 0.91 $\pm$ 0.10   & \tb{0.96 $\pm$ 0.08} \\ \hline
                 14 & 0.87 $\pm$ 0.05       & 0.91 $\pm$ 0.05         & 0.90 $\pm$ 0.08   & \tb{0.92 $\pm$ 0.08} \\ \hline
                 19 & \tb{0.89 $\pm$ 0.04}  & \tb{0.89 $\pm$ 0.04}    & 0.88 $\pm$ 0.06   & 0.88 $\pm$ 0.05 \\ \hline
    \end{tabular}
    \end{adjustbox}
    \caption{P@$\tau$ on the Ionosphere dataset. $\tau\in[0, 1]$ is the predicted positive rate. Results are expressed as mean $\pm$ standard deviation. The columns correspond to logistic regression, Accuracy at the top, the quantile method with one initialization, and the quantile method with three initializations, respectively.}
    \label{tab:ionosphere}
\end{table}

\begin{table}[!htp]
    \centering
    \begin{adjustbox}{width=0.47\textwidth}
    \begin{tabular}{|c|c|c|c|c|}\hline
       $\tau$ (\%)  & LR                & AATP              & Q1                & Q3 \\ \hline
                 1  & 0.26 $\pm$ 0.44   & 0.2 $\pm$ 0.27    & 0.4 $\pm$ 0.49    & \tb{0.43 $\pm$ 0.50} \\ \hline
                 2  & 0.12 $\pm$ 0.19   & 0.23 $\pm$ 0.10   & 0.23 $\pm$0.23    & \tb{0.28 $\pm$ 0.23} \\ \hline
                 3  & 0.09 $\pm$ 0.10   & 0.20 $\pm$ 0.12   & 0.18 $\pm$0.17    & \tb{0.25 $\pm$ 0.16} \\ \hline
                 4  & 0.09 $\pm$ 0.10   & 0.19 $\pm$ 0.13   & 0.16 $\pm$0.14    & \tb{0.23 $\pm$ 0.14} \\ \hline
                 5  & 0.11 $\pm$ 0.09   & 0.17 $\pm$ 0.07   & 0.14 $\pm$0.12    & \tb{0.21 $\pm$ 0.13} \\ \hline
                 6  & 0.11 $\pm$ 0.08   & 0.14 $\pm$ 0.05   & 0.13 $\pm$0.12    & \tb{0.18 $\pm$ 0.10} \\ \hline
    \end{tabular}
    \end{adjustbox}
    \caption{P@$\tau$ on the Housing dataset. $\tau\in[0, 1]$ is the predicted positive rate. The columns correspond to logistic regression, Accuracy at the top, the quantile method with one initialization, and the quantile method with three initializations, respectively.}
    \label{tab:housing}
\end{table}

\subsubsection{KDD Cup 2008}
SVMPerf \cite{joachims2005support} is a standard baseline for methods targeting Precision@K. We compare to the $l^\text{avg}_\text{prec@k}$ surrogate from \cite{kar2015surrogate}, which resolves theoretical issues which arise when applying the structured SVM method to Precision@K. Results are presented in terms of Precision@$\tau$. For this dataset, we consider the loss (\ref{eq:unconstrained_tp_at_ppr}).

\begin{figure}[ht]
\centering
{\includegraphics[width=80mm]{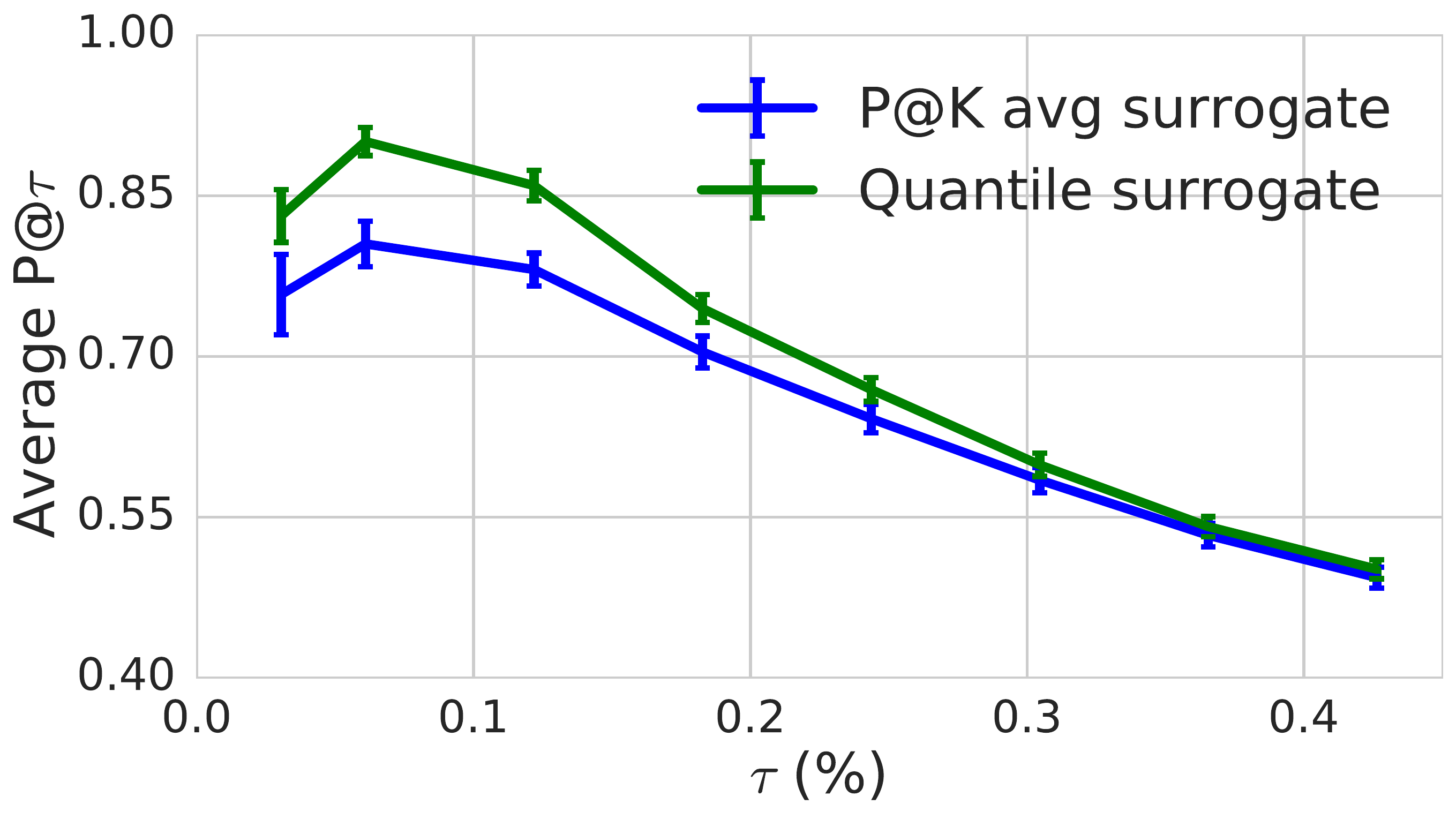}}\\
\caption{P@$\tau$ on the KDD Cup 2008 dataset. Error bars denote 95\% confidence intervals.} \label{fig:kdd}
\end{figure}

Figure \ref{fig:kdd} shows results averaged across 100 random train/test splits of the dataset, with 70\% used for training and the rest reserved for testing. Models with the $l^\text{avg}_\text{prec@k}$ and quantile surrogate losses were evaluated at the same value of $\tau$ for which they were trained, and were learned on the full training set to give the strongest results. The model with the quantile surrogate was trained using stochastic gradient descent with momentum on minibatches of size 1000 for 3 epochs, with randomly initialized parameters.

\subsection{Precision at a fixed recall}
In this section, we evaluate the quantile framework on the target metric of precision at fixed recall for a range of recall values on a variety of datasets. 
\subsubsection{Synthetic dataset}
\begin{figure}[ht]
\centering
{\includegraphics[width=80mm]{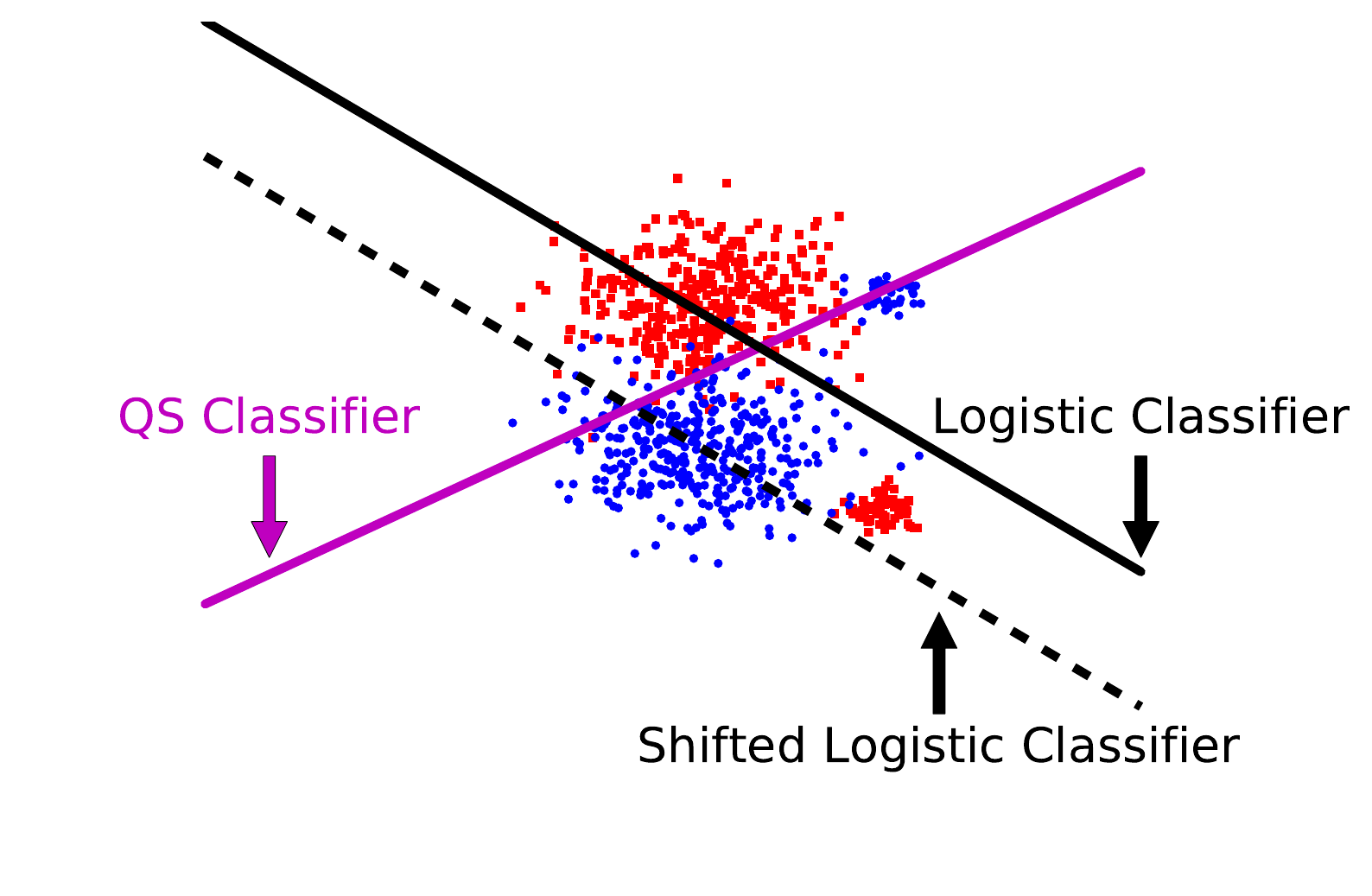}}\\
\caption{Logistic (black) and quantile surrogate loss (magenta) classifiers on the synthetic dataset. The black line depicts the learned threshold of the logistic classifier, while the black-dashed line is the adjusted threshold to satisfy a $90\%$ recall  constraint. The precision of the logistic classifier at recall 0.9 is 0.13, the QS loss classifiers achieves a precision of 0.37.} \label{fig:synthetic}
\end{figure}

To illustrate the benefits of minimizing the precision at recall quantile loss (\ref{eq:unconstrained_fp_at_r}) rather than  maximizing likelihood and adjusting the decision threshold, we consider the synthetic dataset in Figure \ref{fig:synthetic}. The data consists of a mixture of isotropic Gaussians with a prior of $p(y=1) = 0.1$, and the objective is to maximize precision at a recall of 0.95. A similar synthetic dataset is considered in \cite{eban2017scalable}. We initialize the weights of a linear classifier randomly, and minimize (\ref{eq:unconstrained_fp_at_r}) using the Gaussian kernel quantile estimator with $h = 0.05$. For this simple problem, we were unable to find initializations that led to different results.

Adjusting the threshold learned by logistic regression to satisfy the recall constraint results in a classifier which misclassifies most negatives. In contrast, the linear classifier trained using the precision at recall loss performs nearly as well as possible. Threshold adjustment performs poorly in this case because the logistic regression classifier is poorly calibrated; the conditional class probability $p(y=1|x)$ is inaccurate. 

\section{Generalization Bounds}
Consider an $L$-estimator which we call the \textit{interval quantile estimator}
\begin{equation}\label{eq:interval-estimator}
 \hat{q}_I(S, c) = \left|\frac{1}{\floor*{N(k_2 - k_1)}} \sum_{i=\floor*{Nk_1}}^{\floor*{Nk_2}} s_{(i)}\right|,
\end{equation}
where $0 < k_1 < c < k_2$, and $0\leq s_1, \dots, \leq s_N$. This is a generalized version of the upper bound estimator in Eq.(\ref{eq:convex-estimator}), since instead of taking all scores lower than the quantile, we take the average on an arbitrary interval.

We prove a generalization bound for the quantile loss function where the quantile estimator is either the interval estimator or a kernel quantile estimator (Eq.\ref{eq:kernel-estimator}) where $\phi_h$ is bounded and Lipschitz. Moreover, the bound is uniform with respect to model parameters if the model is linear. The conclusion holds for quantile estimators on an arbitrary subset of the feature set $X$, and in particular includes the $P@R$ and $P@k$ case used in the experiments. The proof is similar to that in \cite{kar2015surrogate}, and follows mainly from repeated applications of Hoeffding's inequality (which also holds in the case of sampling without replacement \cite{bardenet2015concentration}).

All proofs are presented in the Appendix.

\begin{proposition}[Uniform Convergence]
\label{prop:unif-conv}
Let $Z = \{(x_1, y_1), \ldots, (x_N, y_N)\}$ be a set of feature-label pairs, $\hat{Z}$ be a sample of $b$ elements chosen from $Z$ (either i.i.d or without replacement), $A$ be some subset of $Z$, $\hat{A} = A \cap \hat{Z}$, and $f(x) = w^T x$ be a linear model with parameters $w$ satisfying $\|w\| \leq C$. Let $\hat{q}(f(A))$, $\hat{q}(f(\hat{A}))$ be either the kernel estimator (\ref{eq:kernel-estimator}) with $\phi$ bounded and Lipschitz, or the interval estimator (\ref{eq:interval-estimator}). Define $L(w;Z, A) = \sum_i (1-y_i) l(f(x_i) - \hat{q}(f(A)))$. Then with probability at least $1 - \delta$,
\begin{equation*}
\begin{split}
 \left|L_q(w; Z, A) - L_q(w; \hat{Z}, \hat{A})\right|= O\left(\sqrt{\frac{1}{b} \log\frac{1}{\delta}}\right)
\end{split}
\end{equation*}
uniformly in $w$.
\end{proposition}

Proposition \ref{prop:unif-conv} gives a uniform bound on the population loss based on the sample loss. However, because the quantile surrogate does not decompose across datapoints, minibatches give biased estimates and the efficacy of stochastic gradient methods is not immediately clear. In the case when $q$ is the lower bound quantile estimator defined in Eq.\ref{eq:convex-estimator}, we have the following convergence bound for stochastic gradient descent on the quantile surrogate loss.

\begin{proposition}
Let $w^*$ be the parameters of a linear model learned by $T$ steps of stochastic gradient descent with batch size $b$ on the quantile loss $L_q$, where $q$ is the lower bound estimator defined in Eq.\ref{eq:convex-estimator}. Then for any parameters $w$, with probability at least $1 - \delta$
\begin{equation*}
\begin{split}
L_q(w^*; Z) \leq L_q(w; Z) + O\left(\sqrt{\frac{1}{b}\log\frac{T}{\delta}}\right) + O\left(\sqrt{\frac{1}{T}}\right).
\end{split}
\end{equation*}
\end{proposition}

The proof of Proposition \ref{prop:unif-conv} depends on the following concentration bounds for the kernel and interval quantile estimators.

\begin{lemma}\label{lemma:stability-kernel-estimators}
Let $F = \{f_1, \dots, f_N\}$ be real numbers sorted in ascending order, and $|f_i| \leq 1$. Let $\hat{F} = \{\hat{f}_1, \dots, \hat{f}_b\}$ be a sample (either i.i.d or without replacement) from the set $F$, also sorted in ascending order. Let $\hat{q}^h(F, c)$, $\hat{q}^h(\hat{F}, c)$ be the kernel quantile estimator defined in Eq.\ref{eq:kernel-estimator}, and assume the kernel function $\phi_h$ is bounded and Lipschitz continuous, then the following holds 
\begin{equation*}
\begin{split}
    &\left|\hat{q}^h(F, c)  - \hat{q}^h(\hat{F}, c)\right| = O\left(\sqrt{\frac{1}{b} \log\frac{1}{\delta}}\right)
\end{split}
\end{equation*}
with probability at least $1 - \delta$. 
\end{lemma}

\begin{lemma}\label{lemma:stability}
Let $F = \{f_1, \dots, f_N\}$ be real numbers sorted in ascending order, and $|f_i| \leq 1$. Let $\{\hat{f}_1, \dots, \hat{f}_b\}$ be a sample (either i.i.d or without replacement) from the set $F$, also sorted in ascending order. Let $0 < k_1< k_2 < 1$. Then the following holds with probability at least $1 - \delta$:
\begin{equation*}
\begin{split}
    &\left|\frac{1}{\floor*{N(k_2 - k_1)}} \sum_{i=\floor*{Nk_1}}^{\floor*{Nk_2}} f_i - \frac{1}{\floor*{b(k_2 - k_1)}} \sum_{j=\floor*{bk_1}}^{\floor*{bk_2}} \hat{f}_j\right| \\
    &= O\left(\sqrt{\frac{1}{b} \log\frac{1}{\delta}}\right).
\end{split}
\end{equation*}
\end{lemma}

\section{Conclusion}
A number of recent papers have demonstrated the value of constrained optimization for machine learning with non-accuracy objectives. We show that for a large class of these problems, the constraint can be eliminated by substitution using the quantile function, which can be effectively estimated in a way which is amenable to standard learning methods. The algorithm enjoys standard empirical risk bounds and strong performance relative to other methods and surrogates for enforcing constraints relevant to non-accuracy objectives. 

While we have presented examples for Precision at a fixed recall and Precision at K, our main contribution -- the use of quantile function estimators in constrained machine learning problems -- is more general.

\section*{Acknowledgements}
The authors would like to thank Ofer Meshi for helpful discussions and insightful comments on the manuscript.





%% file: proofs.tex
\section{Proofs}

\begin{proposition}[Uniform Convergence]
Let $Z = \{(x_1, y_1), \ldots, (x_N, y_N)\}$ be a set of feature-label pairs, $\hat{Z}$ be a sample of $b$ elements chosen from $Z$ (either i.i.d or without replacement), $A$ be some subset of $Z$, $\hat{A} = A \cap \hat{Z}$, and $f(x) = w^T x$ be a linear model with parameters $w$ satisfying $\|w\| \leq C$. Let $\hat{q}(f(A))$, $\hat{q}(f(\hat{A}))$ be either the kernel estimator (\ref{eq:kernel-estimator}) 
with $\phi$ bounded and Lipschitz, or the interval estimator (\ref{eq:interval-estimator}). Define $L(w;Z, A) = \sum_i (1-y_i) l(f(x_i) - \hat{q}(f(A)))$. Then with probability at least $1 - \delta$,
\begin{equation}\label{eq:generalization-bound-1}
\begin{split}
 \left|L_q(w; Z, A) - L_q(w; \hat{Z}, \hat{A})\right|= O\left(\sqrt{\frac{1}{b} \log\frac{1}{\delta}}\right)
\end{split}
\end{equation}
uniformly in $w$.
\end{proposition}
\begin{proof}

We first show pointwise convergence with high probability. Let $\thetahat_N:=\hat{q}(f(A))$, and $\thetahat_b:=\hat{q}(f(\hat{A}))$. We split Eq.(\ref{eq:generalization-bound-1}) to two parts. Let $\mL(x):=l(x)$, we have
\begin{equation*}
\begin{split}
& |L_q(w; Z, A) - L_q(w; \hat{Z}, \hat{A})|\\
& \leq \left|\frac{1}{N}\sum_{i = 1}^N \mL(f(x_i) - \thetahat_N)-\frac{1}{b}\sum_{i = 1}^b \mL(f(\xh_i) - \thetahat_N) \right|  \quad \text{(A)}\\
& + \left| \frac{1}{b}\sum_{i = 1}^b \mL(f(x_i) - \thetahat_N)
 - \frac{1}{b}\sum_{i = 1}^b\mL(f(x_i) - \thetahat_b)\right|  \\
&\leq (A) + \left|\thetahat_N - \thetahat_b\right| 
\end{split}
\end{equation*}
where the last line follows from the logloss being Lipschitz with constant $1$. 
Since $w$ is bounded, the scores $f(x_i)$ are bounded and an application of Hoeffding's inequality gives $\text{(A)}=O\left(\frac{1}{b} \log\frac{1}{\delta}\right)$ w.h.p. 
The bound for the term $|\thetahat_N - \thetahat_b| $ is proved in Lemma \ref{lemma:stability} and Lemma \ref{lemma:stability-kernel-estimators} by setting $f_{i'} = f(x_i), i\in A$, and scaling $f_i$ such that $|f_i| < 1$.

Because $f$ is linear, it can be shown that the bound is uniform with respect to $w$ by arguing along the lines of Theorem 12 in \cite{kar2015surrogate}. Namely, we first show that the quantile loss is Lipschitz with respect to model parameters $w$. Note that logloss is Lipschitz continuous with constant 1, and $\theta_N(w)$, $\theta_b(w)$ are Lipschitz with constant $\max_i\{\|x_i\|_\infty\}$. Since the quantile loss $L_q$ is a composition of logloss and the quantile estimator, $L_q$ is Lipschitz. Applying the $L_\infty$ covering number argument in \cite{zhang2002covering} for linear models, together with pointwise convergence, gives uniform convergence. 
\end{proof}

\begin{proposition}
Let $w^*$ be the parameters of a linear model learned by $T$ steps of stochastic gradient descent with batch size $b$ on the quantile loss $L_q$, where $q$ is the lower bound estimator defined in Eq.\ref{eq:convex-estimator}. Then for any parameters $w$, with probability at least $1 - \delta$
\begin{equation}
\label{eq:data-stream}
\begin{split}
L_q(w^*; Z) \leq L_q(w; Z) + O\left(\sqrt{\frac{1}{b}\log\frac{T}{\delta}}\right) + O\left(\sqrt{\frac{1}{T}}\right).
\end{split}
\end{equation}
\end{proposition}

\begin{proof}
Since the quantile loss $L_q$ with the lower bound quantile estimator is convex, we can apply Theorem 6 from \cite{kar2014online} which states that any convex loss satisfying the uniform convergence property in Proposition \ref{prop:unif-conv} satisfies Eq.\ref{eq:data-stream}. 
\end{proof}

\begin{lemma}
Let $F = \{f_1, \dots, f_N\}$ be real numbers sorted in ascending order, and $|f_i| \leq 1$. Let $\hat{F} = \{\hat{f}_1, \dots, \hat{f}_b\}$ be a sample (either i.i.d or without replacement) from the set $F$, also sorted in ascending order. Let $\hat{q}^h(F, c)$, $\hat{q}^h(\hat{F}, c)$ be the kernel quantile estimator defined in Eq.\ref{eq:kernel-estimator}, and assume the kernel function $\phi_h$ is bounded and Lipschitz continuous, then the following holds 
\begin{equation*}
\begin{split}
    &\left|\hat{q}^h(F, c)  - \hat{q}^h(\hat{F}, c)\right| = O\left(\sqrt{\frac{1}{b} \log\frac{1}{\delta}}\right)
\end{split}
\end{equation*}
with probability at least $1 - \delta$. 
\end{lemma}

\begin{proof}
For simplicity, assume the numbers $f_i, i=1\dots N$ are distinct. Define $p(x):=\frac{|\{\hat{f_j} \leq x\}|}{b}$, and writing $p_k = p(f_k)$ for short, we have 
\begin{equation}
\label{eq:lemma-kernel}
\begin{split}
    &\hat{q}^h(\hat{F}, c) := \frac{1}{b} \sum_{i=1}^{b} \phi_h\left(\frac{i^*}{b} - c\right) \hat{f}_i\\
    &= \frac{1}{b} \sum_{i=1}^{b} \phi_h\left(p(\hat{f}_i) - c \right) \hat{f}_i\\
    &= \frac{1}{b} \sum_{k=1}^N \phi_h(p(f_k) - c) f_k \times |\{\hat{f}_i = f_k\}|  \\
    &= \frac{1}{b} \sum_{k=1}^N \phi_h(p(f_k) - c) f_k \times \left(|\{\hat{f}_i \leq f_k\}| - |\{\hat{f}_i \leq f_{k - 1}\}|\right)\\
    &= \sum_{k=1}^{N} \phi_h\left(p_k - c\right) \left(p_k - p_{k-1}\right)f_k,
\end{split}
\end{equation}
with $p_0 :=0$. Since $|p_k - \frac{k}{N}| = |\frac{1}{b}\sum_{i=1}^b \one_{\hat{f}_i \leq f_k} - \bbE\one_{\hat{f}_i \leq f_k}|$, we have for all $k$, $|p_k - \frac{k}{N}| = O\left(\sqrt{\frac{1}{b} \log\frac{1}{\delta}}\right)$ w.h.p. by Hoeffding's inequality. Similarly, we also have $|p_k - p_{k-1} - \frac{1}{N}| = O\left(\sqrt{\frac{1}{b} \log\frac{1}{\delta}}\right).$ Let $|\phi_h|$ be bounded by $M$ with Lipschitz constant $K$. Then by Eq.\ref{eq:lemma-kernel}, 
\begin{equation*}
\begin{split}
    &\hat{q}^h(\hat{F}, c) = \sum_{k=1}^{N} \phi_h(p_k - c) (p_k - p_{k-1})f_k \\
    &= \sum_{k=1}^{N} \frac{1}{N}\phi_h\left(\frac{k}{N} - c\right) f_k + (K+M)O\left(\sqrt{\frac{1}{b} \log\frac{1}{\delta}}\right)\\
    &= \hat{q}^h(F, c) + O\left(\sqrt{\frac{1}{b} \log\frac{1}{\delta}}\right)
\end{split}
\end{equation*}
holds w.h.p. The case where $f_i$ are not distinct could be proved similarly by considering a non-uniform draw from the set of unique elements in $\{f_i\}$. 
\end{proof}

\begin{lemma}
Let $F = \{f_1, \dots, f_N\}$ be real numbers sorted in ascending order, and $|f_i| \leq 1$. Let $\{\hat{f}_1, \dots, \hat{f}_b\}$ be a sample (either i.i.d or without replacement) from the set $F$, also sorted in ascending order. Let $0 < k_1< k_2 < 1$. Then the following holds with probability at least $1 - \delta$:
\begin{equation*}
\begin{split}
    &\left|\frac{1}{\floor*{N(k_2 - k_1)}} \sum_{i=\floor*{Nk_1}}^{\floor*{Nk_2}} f_i - \frac{1}{\floor*{b(k_2 - k_1)}} \sum_{j=\floor*{bk_1}}^{\floor*{bk_2}} \hat{f}_j\right| \\
    &= O\left(\sqrt{\frac{1}{b} \log\frac{1}{\delta}}\right).
\end{split}
\end{equation*}
\end{lemma}

\begin{proof}
For convenience, assume that $Nk_1$, $Nk_2$, $bk_1$, and $bk_2$ are all integers. Let $w_i = f_{\floor*{Nk_i}}$, $\hat{w}_i = \hat{f}_{\floor*{b k_i}}$ for $i = 1, 2$. Define $\Tc_1(f) = \mathbbm{1}_{f > w_1}$, $\Tc_2(f) = \mathbbm{1}_{f \leq w_2}$, and $\Tch_1(f) = \mathbbm{1}_{f > \hat{w}_1}$, $\Tch_2(f) = \mathbbm{1}_{f \leq \hat{w}_2}$. Letting $c = \frac{1}{k_2 - k_1}$, we have the following estimate:
\begin{equation*}
\begin{split}
    &\left|\frac{1}{\floor*{N(k_2 - k_1)}} \sum_{i=\floor*{Nk_1+1}}^{\floor*{Nk_2}} f_i - \frac{1}{\floor*{b(k_2 - k_1)}} \sum_{j=\floor*{bk_1}+1}^{\floor*{bk_2}} \hat{f}_j\right| 
    \\&= c\left|\frac{1}{N}\sum_{i=1}^N\Tc_1(f_i)\Tc_2(f_i)f_i - \frac{1}{b}\sum_{i=1}^b\Tch_1(\hat{f}_i)\Tch_2(\hat{f}_i)\hat{f}_i \right| \\
    &\leq c\left|\frac{1}{N}\sum_{i=1}^N\Tc_1(f_i)\Tc_2(f_i)f_i - \frac{1}{b}\sum_{i=1}^b\Tc_1(\hat{f}_i)\Tc_2(\hat{f}_i)\hat{f}_i \right| \text{(A)}\\
    &+ c \left|\frac{1}{b}\sum_{i=1}^b\Tc_1(\hat{f}_i)\Tc_2(\hat{f}_i)\hat{f}_i - \frac{1}{b}\sum_{i=1}^b\Tch_1(\hat{f}_i)\Tch_2(\hat{f}_i)\hat{f}_i\right|. \text{\hspace{3pt}(B)}
\end{split}
\end{equation*}
(A) is $O\left(\sqrt{\frac{1}{b} \log\frac{1}{\delta}}\right)$ w.h.p. by Hoeffding's inequality (since $w_i$ is fixed with respect to the samples $\hat{f}_i$). For (B), we use the observation that
\[
\begin{split}
|&\Tc_1(x)\Tc_2(x) - \Tch_1(x)\Tch_2(x)|  = |\Tc_1(x) - \Tch_1(x) \\&+ \Tc_2(x) - \Tch_2(x)|
\end{split}
\]
holds for all $w_i$, $\hat{w}_i$, and $x$, which can be verified by checking all cases. Using the equality above, we have

\begin{equation*}
\begin{split}
&\left|\frac{1}{b}\left(\sum_{i=1}^b\Tc_1(\hat{f}_i)\Tc_2(\hat{f}_i) - \sum_{i=1}^b\Tch_1(\hat{f}_i)\Tch_2(\hat{f}_i)\right)\hat{f}_i\right| \\
&\leq \left|\frac{1}{b}\left(\sum_{i=1}^b\Tc_1(\hat{f}_i)\Tc_2(\hat{f}_i) - \sum_{i=1}^b\Tch_1(\hat{f}_i)\Tch_2(\hat{f}_i)\right)\right|\\
 &\leq \frac{1}{b}\sum_{i=1}^b\left|\left(\Tc_1(\hat{f}_i) - \Tch_1(\hat{f}_i)\right) + \left(\Tc_2(\hat{f}_i) - \Tch_2(\hat{f}_i)\right)\right| \\
 &\leq  \frac{1}{b}\sum_{i=1}^b\left|\left(\Tc_1(\hat{f}_i) - \Tch_1(\hat{f}_i)\right) \right| +\frac{1}{b}\sum_{i=1}^b\left|\left(\Tc_2(\hat{f}_i) - \Tch_2(\hat{f}_i)\right) \right|\\
 &=  \left|\frac{1}{b}\sum_{i=1}^b\left(\Tc_1(\hat{f}_i) - \Tch_1(\hat{f}_i)\right) \right| +\left|\frac{1}{b}\sum_{i=1}^b\left(\Tc_2(\hat{f}_i) - \Tch_2(\hat{f}_i)\right) \right|,
\end{split}
\end{equation*}
where the last line follows from the fact that the function $\Tc_i(x)- \Tch_i(x)$ has a fixed sign for all $x$. Note that $$\left|\frac{1}{b}\sum_{i=1}^b\left(\Tc_1(\hat{v}_i) - \Tch_1(\hat{f}_i)\right) \right| = \left|\frac{1}{ b}\sum_{i=1}^b\Tc_1(\hat{f}_i) - k_1\right|,$$ which is again bounded by $O\left(\sqrt{\frac{1}{b} \log\frac{1}{\delta}}\right)$ w.h.p. via Hoeffding's inequality. Arguing similarly for the other term with $\Tc_2$ and $\Tch_2$, we conclude the proof of the lemma.
\end{proof}

%% file: arxiv_submission.bbl
\begin{thebibliography}{10}

\bibitem{bardenet2015concentration}
R{\'e}mi Bardenet, Odalric-Ambrym Maillard, et~al.
\newblock Concentration inequalities for sampling without replacement.
\newblock {\em Bernoulli}, 21(3):1361--1385, 2015.

\bibitem{boyd2012accuracy}
Stephen Boyd, Corinna Cortes, Mehryar Mohri, and Ana Radovanovic.
\newblock Accuracy at the top.
\newblock In {\em Advances in neural information processing systems}, pages
  953--961, 2012.

\bibitem{boyd2004convex}
Stephen Boyd and Lieven Vandenberghe.
\newblock {\em Convex optimization}.
\newblock Cambridge university press, 2004.

\bibitem{cheng1995bernstein}
Cheng Cheng.
\newblock The bernstein polynomial estimator of a smooth quantile function.
\newblock {\em Statistics \& probability letters}, 24(4):321--330, 1995.

\bibitem{davenport2010tuning}
Mark~A Davenport, Richard~G Baraniuk, and Clayton~D Scott.
\newblock Tuning support vector machines for minimax and neyman-pearson
  classification.
\newblock {\em IEEE Transactions on Pattern Analysis and Machine Intelligence},
  32(10):1888--1898, 2010.

\bibitem{eban2017scalable}
Elad Eban, Mariano Schain, Alan Mackey, Ariel Gordon, Ryan Rifkin, and Gal
  Elidan.
\newblock Scalable learning of non-decomposable objectives.
\newblock In {\em Artificial Intelligence and Statistics}, pages 832--840,
  2017.

\bibitem{elkan2001foundations}
Charles Elkan.
\newblock The foundations of cost-sensitive learning.
\newblock In {\em International joint conference on artificial intelligence},
  volume~17, pages 973--978. Lawrence Erlbaum Associates Ltd, 2001.

\bibitem{goh2016satisfying}
Gabriel Goh, Andrew Cotter, Maya Gupta, and Michael~P Friedlander.
\newblock Satisfying real-world goals with dataset constraints.
\newblock In {\em Advances in Neural Information Processing Systems}, pages
  2415--2423, 2016.

\bibitem{harrell1982new}
Frank~E Harrell and CE~Davis.
\newblock A new distribution-free quantile estimator.
\newblock {\em Biometrika}, 69(3):635--640, 1982.

\bibitem{he2009learning}
Haibo He and Edwardo~A Garcia.
\newblock Learning from imbalanced data.
\newblock {\em IEEE Transactions on knowledge and data engineering},
  21(9):1263--1284, 2009.

\bibitem{joachims2005support}
Thorsten Joachims.
\newblock A support vector method for multivariate performance measures.
\newblock In {\em Proceedings of the 22nd international conference on Machine
  learning}, pages 377--384. ACM, 2005.

\bibitem{kalgh1982generalized}
WD~Kalgh and Peter~A Lachenbruch.
\newblock A generalized quantile estimator.
\newblock {\em Communications in Statistics-Theory and Methods},
  11(19):2217--2238, 1982.

\bibitem{kar2014online}
Purushottam Kar, Harikrishna Narasimhan, and Prateek Jain.
\newblock Online and stochastic gradient methods for non-decomposable loss
  functions.
\newblock In {\em Advances in Neural Information Processing Systems}, pages
  694--702, 2014.

\bibitem{kar2015surrogate}
Purushottam Kar, Harikrishna Narasimhan, and Prateek Jain.
\newblock Surrogate functions for maximizing precision at the top.
\newblock In {\em International Conference on Machine Learning}, pages
  189--198, 2015.

\bibitem{koller2009probabilistic}
Daphne Koller and Nir Friedman.
\newblock {\em Probabilistic graphical models: principles and techniques}.
\newblock MIT press, 2009.

\bibitem{koyejo2014consistent}
Oluwasanmi~O Koyejo, Nagarajan Natarajan, Pradeep~K Ravikumar, and Inderjit~S
  Dhillon.
\newblock Consistent binary classification with generalized performance
  metrics.
\newblock In {\em Advances in Neural Information Processing Systems}, pages
  2744--2752, 2014.

\bibitem{kull2017beta}
Meelis Kull, Telmo Silva~Filho, and Peter Flach.
\newblock Beta calibration: a well-founded and easily implemented improvement
  on logistic calibration for binary classifiers.
\newblock In {\em Artificial Intelligence and Statistics}, pages 623--631,
  2017.

\bibitem{margineantu2002class}
Dragos~D Margineantu.
\newblock Class probability estimation and cost-sensitive classification
  decisions.
\newblock In {\em ECML}, pages 270--281. Springer, 2002.

\bibitem{pepelyshev2014estimation}
Andrey Pepelyshev, E~Rafaj{\l}owicz, and A~Steland.
\newblock Estimation of the quantile function using bernstein--durrmeyer
  polynomials.
\newblock {\em Journal of Nonparametric Statistics}, 26(1):1--20, 2014.

\bibitem{sheather1990kernel}
Simon~J Sheather and James~Stephen Marron.
\newblock Kernel quantile estimators.
\newblock {\em Journal of the American Statistical Association},
  85(410):410--416, 1990.

\bibitem{taskar2004max}
Ben Taskar, Carlos Guestrin, and Daphne Koller.
\newblock Max-margin markov networks.
\newblock In {\em Advances in neural information processing systems}, pages
  25--32, 2004.

\bibitem{tong2016survey}
Xin Tong, Yang Feng, and Anqi Zhao.
\newblock A survey on neyman-pearson classification and suggestions for future
  research.
\newblock {\em Wiley Interdisciplinary Reviews: Computational Statistics},
  8(2):64--81, 2016.

\bibitem{tsochantaridis2005large}
Ioannis Tsochantaridis, Thorsten Joachims, Thomas Hofmann, and Yasemin Altun.
\newblock Large margin methods for structured and interdependent output
  variables.
\newblock {\em Journal of machine learning research}, 6(Sep):1453--1484, 2005.

\bibitem{zhang2002covering}
Tong Zhang.
\newblock Covering number bounds of certain regularized linear function
  classes.
\newblock {\em Journal of Machine Learning Research}, 2(Mar):527--550, 2002.

\bibitem{zielinski2004optimal}
Ryszard Zielinski.
\newblock {\em Optimal Quantile Estimators: Small Sample Approach}.
\newblock Polish Academy of Sciences. Institute of Mathematics, 2004.

\end{thebibliography}
